\documentclass{article}

\usepackage[preprint]{neurips}

\usepackage{natbib}
\setcitestyle{numbers,square}
\usepackage[utf8]{inputenc} % allow utf-8 input
\usepackage[T1]{fontenc}    % use 8-bit T1 fonts
\usepackage{hyperref}       % hyperlinks
\usepackage{url}            % simple URL typesetting
\usepackage{booktabs}       % professional-quality tables
\usepackage{amsfonts}       % blackboard math symbols
\usepackage{nicefrac}       % compact symbols for 1/2, etc.
\usepackage{microtype}      % microtypography
\usepackage{overpic}         % colors
\usepackage{multirow}
\usepackage[dvipsnames,table,xcdraw]{xcolor}

\usepackage{pifont}

\newcommand{\figref}[1]{Fig.~\ref{#1}}
\newcommand{\tabref}[1]{Tab.~\ref{#1}}

\def\ie{\emph{i.e.}}

\def\etal{{\em et al.~}}

\title{Segment Anything is A Good Pseudo-label Generator for Weakly Supervised Semantic Segmentation}

\author{
Peng-Tao Jiang\thanks{* denotes equal contribution.} \\ State Key Lab of CAD\&CG, Zhejiang University
\And Yuqi Yang* \\ TMCC, CS, Nankai University %   % \\ \texttt{email} \\
}

\begin{document}

\maketitle

\begin{abstract}
Weakly supervised semantic segmentation with weak labels 
is a long-lived ill-posed problem. 
Mainstream methods mainly focus on improving the quality of 
pseudo labels.
In this report, we attempt to explore the potential of 'prompt to masks' 
from the powerful class-agnostic large segmentation model, \ie, 
segment-anything.
Specifically, different weak labels are used as prompts to 
the segment-anything model, generating precise class masks.
The class masks are utilized to generate pseudo labels 
to train the segmentation networks.
We have conducted extensive experiments on PASCAL VOC 2012 
dataset.
Experiments demonstrate that segment-anything can serve 
as a good pseudo-label generator.
The code will be made publicly available.
\end{abstract}

\section{Introduction}
Semantic segmentation \cite{long2015fully,zhao2016pyramid,chen2017deeplab} 
is a classic computer vision task that aims to classify each pixel in the image.
Training segmentation models usually needs 
large-scale finely-annotated segmentation datasets, such as PASCAL VOC \cite{everingham2015pascal}, 
MS COCO \cite{lin2014microsoft}, ADE20K \cite{zhou2017scene}.  
However, constructing such large-scale datasets consumes much 
time and cost, even using polygon annotations.
Thus, in recent years, researchers have attempted to focus 
on weakly supervised semantic segmentation that aims to utilize cheaper 
annotations than pixel-level annotations to train segmentation 
models.
The cheaper annotations include image labels \cite{huang2018weakly}, 
points \cite{bearman2016s}, scribbles \cite{lin2016scribblesup}, 
and bounding boxes \cite{lee2021bbam}.
Previous mainstream works \cite{khoreva2017simple,jiang2021online,lee2021railroad} 
follow an idea that utilizes the cheaper annotations as initial spatial priors 
to generate pseudo labels \cite{wei2018revisiting} 
or learn affinity propagation \cite{lin2016scribblesup}.

Recently, large models \cite{bommasani2021opportunities,brown2020language,fang2022eva,radford2021learning,oquab2023dinov2} 
have dominated computer vision and natural language processing,
which benefit from large-scale data and billions of model parameters.
A large segmentation model, called segment-anything \cite{kirillov2023segment}, 
is proposed for the segmentation field. 
The segment-anything model (SAM) can receive different kinds of spatial prompts 
and output several object masks, where the spatial prompts include points, 
bounding boxes, and texts.
We observe that object masks usually have precise boundaries, which 
can facilitate the weakly supervised semantic segmentation task.

In this report, we propose to utilize SAM to generate pseudo labels 
and utilize them to train the segmentation networks.
Specifically, we attempt to explore different weak annotations as 
prompts for SAM and generate object masks with precise boundaries.
We present a detailed analysis about the impact of different prompts 
on the quality of pseudo labels.
Finally, we present the final segmentation results of different prompts.
Using scribbles as prompts, we can generate precise pseudo labels
with an 89.7\% mIoU score on PASCAL VOC 2012 train set, 
approximating ground-truth labels.
The final segmentation model achieves a 76.6 \% 
mIoU score on the test set.

\section{Method} 
\begin{figure}[t]
    \centering
    %\footnotesize
    \setlength\tabcolsep{1pt}
    \begin{overpic}[width=\linewidth]{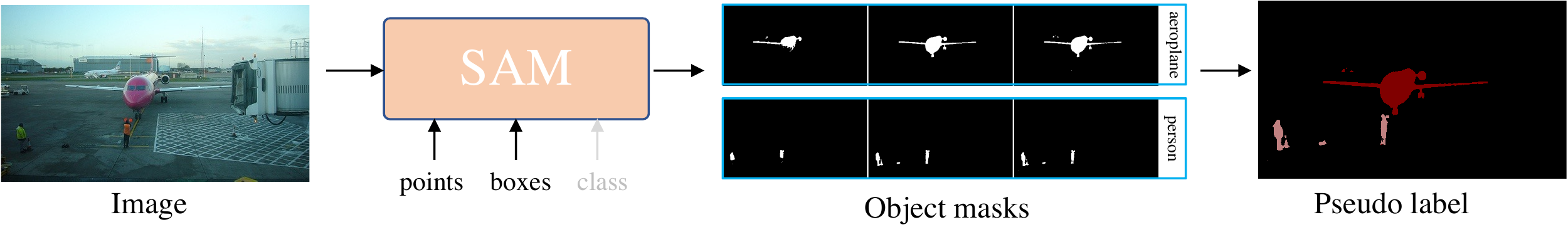}
    \end{overpic}
    \caption{Pipeline of our method. Text prompt is not available in SAM.
    }\label{fig:pipeline}
    % \vspace{-1pt}
\end{figure}

In this section, we introduce how we utilize different weak labels 
to generate prompts for SAM.
The overall pipeline of our method is shown in \figref{fig:pipeline}.
SAM can receive points, boxes, and texts as input prompts 
and output the corresponding object masks located by these prompts.
Note SAM does not make the text prompt available now.

\subsection{Image-level Labels}
Image-level labels only contain information about which category 
exists in an image.
It does not provide any object localization information.
Previous weakly supervised semantic methods \cite{ahn2018learning,papandreou2015weakly,wang2018weakly} 
usually utilize class activation maps (CAMs) \cite{zhou2016learning,selvaraju2017grad,jiang2021layercam} 
to generate pseudo labels.
They mainly focus on improving the localization ability of CAMs.
Early works \cite{wei2017object,hou2018self,jiang2019integral,sun2020mining} 
aim to locate more integral object regions as CAMs usually locate small object regions.
They generate accurate pseudo labels with the aid of saliency maps \cite{liu2019simple} 
that usually have precise object boundaries.
Another line of works \cite{ahn2018learning,ahn2019weakly} exploits 
pixel affinities to locate integral object regions 
with precise object boundaries.

In this report, we aim to explore the potential of large segmentation models, \ie, 
segment-anything, to generate pseudo labels.
We propose two methods to utilize image-level labels with SAM. 
\textbf{(i)} One is to sample points on object regions located by CAMs 
and then utilize the sampled points as prompts.
\textbf{(ii)} Another is to generate object masks for all spatial 
locations first and utilize BLIP-2 \cite{li2023blip} to classify each mask.
In the following, we introduce these two methods in detail.

\begin{table}[b]
    \centering
    \small
    \renewcommand{\arraystretch}{1.0}%{1.05}
    \setlength\tabcolsep{1.0mm}
    \caption{Comparisons of mIoU scores under different settings.
    $\mbox{mIoU}_{train}$ denotes the mIoU score of the pseudo segmentation labels 
    on the training set.
    } 
    \vspace{-5pt}
    \begin{tabular}{c|cccc|c} \toprule[1.0pt]
      Annotations  & All confident pixels &   Sample confident pixels  & Iterative input   & Negative points   & $\mbox{mIoU}_{train}$    \\ 
      \midrule[0.8pt] \multirow{4}*{Image-level labels} 
      &                &              &            &               &    47.1                                                           \\
      &  \checkmark    &              &            &               &    50.9    \\ 
      &                & \checkmark   &            &               &    61.5    \\ 
      &                & \checkmark   & \checkmark &               &    59.4    \\ 
      &                & \checkmark   &            &  \checkmark   & \textbf{61.9} \\ 
      \midrule[0.8pt] \multirow{2}*{Points} 
      &  \checkmark    &              &            &               &    69.2    \\ 
      &  \checkmark    &              & \checkmark &               & \textbf{71.7} \\
      &  \checkmark    &              & \checkmark &   \checkmark  &    71.5    \\    
      \midrule[0.8pt]\multirow{4}*{Scribbles} 
      &  \checkmark    &              &            &               &    74.6    \\ 
      &                & \checkmark   &            &               &    81.0    \\ 
      &                & \checkmark   & \checkmark &               &    84.3    \\ 
      &                & \checkmark   & \checkmark &  \checkmark   & \textbf{89.7} \\ 
      \midrule[0.8pt]\multirow{1}*{Bounding boxes} 
      &  \checkmark    &              &            &               & \textbf{91.5} \\  
   \bottomrule[1.0pt]
    \end{tabular}
    \vspace{-10pt}
    \label{tab:ablation}
  \end{table}

First, we study how to sample points from CAMs to generate pseudo labels.
To generate point prompts, we utilize two settings to sample points from CAMs. 
The first is to utilize all confidence pixels in CAMs as a prompt.
Another is to sample confidence pixels from CAMs as a prompt, 
where the sampled pixels exhibit higher values 
than their neighboring pixels within a given range.
As shown in \tabref{tab:ablation}, we can see that sampling confident 
pixels achieves a higher mIoU score.
Besides, SAM provides a mechanism that iteratively receives 
new point prompts for mask refinement.
It can be seen that the iterative refinement cannot improve the 
quality of pseudo labels. 
We analyze that this is because point prompts located by CAMs 
have much noise, which will harm the refinement.
Finally, when multiple classes exist in the image, they can 
serve as negative point prompts for other classes, bringing 
0.4\% mIoU score improvement.
In this report, we only exploit the basic CAMs to locate 
point prompts.
Better CAMs \cite{wang2020score,wang2020self,wu2021embedded,yao2021non} 
can be further explored.
In \figref{fig:failure}, we present several failure cases 
from point prompts located by CAMs.
In the top row, SAM generates the wrong object masks 
due to the coarse locations by CAMs.
In the second row, SAM locates a part of the dining table 
because multiple objects are placed on the tables.

\newcommand{\addFig}[1]{\includegraphics[width=0.16\linewidth]{figs/failure/#1}}
\newcommand{\addFigs}[1]{\addFig{#1.jpg} & \addFig{#1.png}  & \addFig{#1_gt.png}}

\begin{figure}[t] 
  \centering
  \small
%   \vspace{-8pt}
  \setlength\tabcolsep{0.3pt}
  \renewcommand{\arraystretch}{0.8}
  \begin{tabular}{cccccc}
    \addFigs{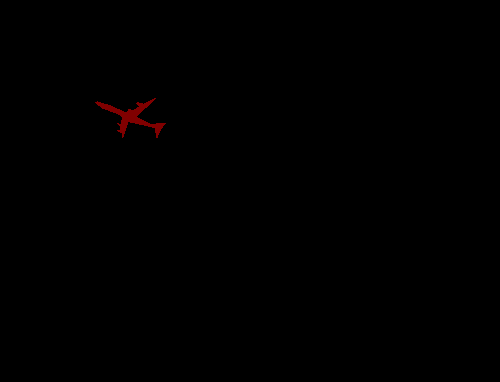} & \addFigs{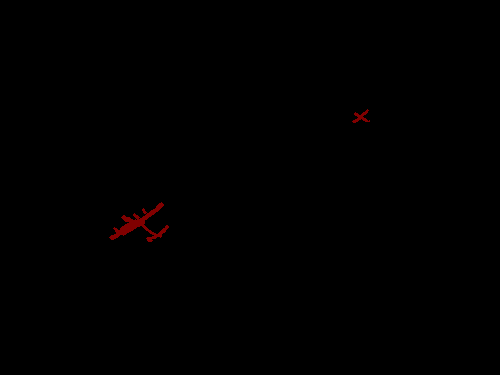}\\
    \addFigs{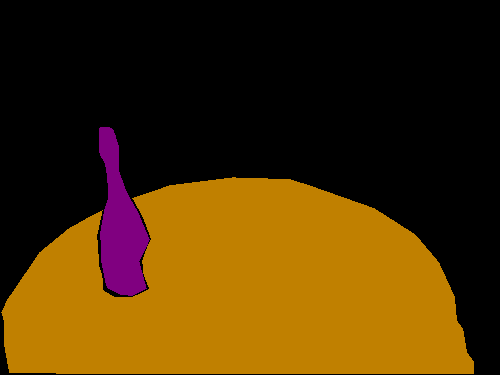} & \addFigs{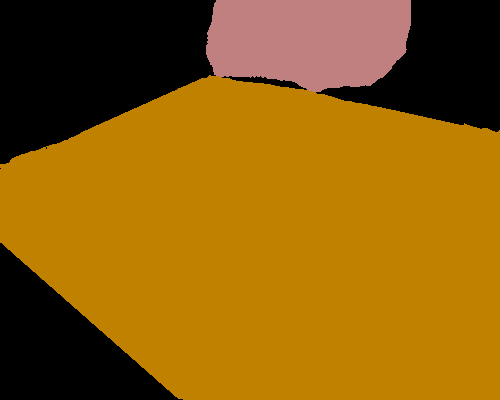} \\
    Image & Pseudo label  & GT  & Image & Pseudo label  & GT  \\
  \end{tabular}
     \vspace{-5pt}
    \caption{Failure pseudo labels generated by SAM with image-level labels.
    }\label{fig:failure}
    \vspace{-15pt}
\end{figure}

Furthermore, we utilize SAM with BLIP-2 \cite{li2023blip} 
to generate pseudo labels.
Specifically, we first generate masks for all strided points.
Then the classification of each mask is restricted 
to the target classes with the background.
The mIoU score of pseudo labels achieves 3.3\% higher 
than AdvCAM \cite{lee2021anti} with the refinement 
of IRNet \cite{ahn2018learning}.
Such surprising performance indicates the effectiveness 
of using SAM to generate pseudo labels.

\subsection{Points}
Point labels \cite{bearman2016s} locate one pixel in each object 
for all target classes, which can directly serve as a prompt to SAM.
For each point, SAM will generate corresponding object masks.
We utilize these masks to compose the final pseudo labels.
As point labels can be regarded as a particular case of scribbles, 
the settings are nearly the same with scribble labels.
As shown in \tabref{tab:ablation}, iteratively inputting each pixel 
of a class achieves the highest mIoU score.

\begin{table}[h]
    \centering
    %\scriptsize
    \small
        \caption{Quantitative comparisons of the pseudo labels 
        of different methods.
        }\label{tab:pseudo_labels}
        \vspace{-5pt}
    \renewcommand{\arraystretch}{1.10}%{1.05}
    \setlength{\tabcolsep}{3.mm}{
    \begin{tabular}{c|c|c|c} \toprule[1pt] 
    Annotations  & Methods                           &   Publication   & Train (\%)  
    \\ \midrule[0.8pt]
    \multirow{5}*{Image-level labels} 
    &   CAM \cite{zhou2016learning,wu2019wider}    & CVPR'16    &  47.1          \\
    &   AdvCAM \cite{lee2021anti}                  & CVPR'21    &  55.6          \\
    &AdvCAM+IRNet \cite{ahn2019weakly}             & CVPR'21    &  69.9          \\
    &   CLIP-ES \cite{lin2022clip}                 &  arXiv'22  &  70.8          \\
    &   CAM + SAM                                  &  --        &  61.9          \\
    &   CLIP-ES + SAM                              &  --        &  72.4          \\
    &   SAM + BLIP-2                               &  --        &  73.2          \\
    \midrule[0.8pt]
    \multirow{1}*{Points}             
    &  Point + SAM                                 &  --        &  69.1           \\
    \midrule[0.8pt]
    \multirow{1}*{Scribbles}                       
    &  Scribbles + SAM                             &            &  89.7           \\
    \midrule[0.8pt]
    \multirow{1}*{Bounding boxes}     
    % &  BBAM \cite{lee2021bbam}                     & CVPR'21    &  --             \\
    &  Bounding boxes + SAM                        &  --        &  91.5           \\ 
    \bottomrule[1pt]
    \end{tabular} }
    \vspace{-5pt}
\end{table}

\subsection{Scribbles}
Scribble labels \cite{wang2019boundary} are a set of pixels 
in each object for all target classes.
Scribbles provide more object localization information 
than image-level labels and points.
Lin \etal \cite{lin2016scribblesup} utilized the graphical model 
to propagate the scribble information to unknown pixels.
Tang \etal \cite{tang2018normalized} designed a normalized cut loss 
to learn segmentation networks based on scribble labels.

We utilize scribble labels as the prompt to SAM.
As scribble labels in each object contain multiple pixels, 
there are several settings to input scribbles to SAM. 
We have conducted experiments for these settings.
As shown in \tabref{tab:pseudo_labels}, we find that 
sampling 20\% scribble pixels outperforms inputting 
all scribble pixels in an object by 6.4\%.
Besides, iteratively inputting scribble pixels of a class 
can further improve the performance by 3.3\%.
We analyze that iterative input is more effective 
for scribbles and points than image-level labels 
due to accurate point locations.
Finally, when inputting the scribble pixels 
of one class, the scribble pixels of other 
classes can be regarded as negative points.
We can see that adding negative points can further 
improve the quality of pseudo labels.
Using the best pseudo labels from scribble prompts, 
DeepLab-v2 \cite{chen2017deeplab} can reach 75.9\% and 76.6\% 
mIoU scores on the validation and test sets, 
as shown in \tabref{tab:comps_pascal_resnet}.

\subsection{Bounding Boxes}
Bounding box labels \cite{dai2015boxsup} provide a tight box 
for each object of a class.
Given bounding box labels, we send each box of a class 
to SAM and generate its corresponding object masks.
As shown in \tabref{tab:pseudo_labels}, it achieves 91.5\% 
mIoU score on the train set, which are the best pseudo labels 
among all weak annotations.
Note we do not add the negative points for bounding 
box prompts as the bounding boxes cannot provide accurate point locations.

\begin{table}[h]
    \centering
    %\scriptsize
    \small
        \caption{Quantitative comparisons of the pseudo labels 
        of different methods.
        }\label{tab:comps_pascal_resnet}
        \vspace{-5pt}
    \renewcommand{\arraystretch}{1.10}%{1.05}
    \setlength{\tabcolsep}{2.mm}{
    \begin{tabular}{c|c|c|c|c} \toprule[1pt] 
    Annotations  & Methods  &   Publication          & Val (\%)   & Test (\%) 
    \\ \midrule[0.8pt]
    \multirow{2}*{Image-level labels} 
     &   AdvCAM \cite{lee2021anti}     & CVPR'21     & 68.1       & 68.0   \\
     &   EPS \cite{lee2021railroad}    & CVPR'22     & 70.9       & 70.8   \\
     & Image-level labels + SAM        &  --         & 71.1       & 72.2   \\
    \midrule[0.8pt]
    \multirow{2}*{Points}            
    & WhatsPoint \cite{bearman2016s}   & ECCV'16     & 46.1       & -      \\
    & Points + SAM                     &  --         & 69.0       & 68.7   \\
    \midrule[0.8pt]
    \multirow{2}*{Scribbles}          
    & ScribbleSup \cite{lin2016scribblesup} & CVPR'16  & 63.1     & -    \\ 
    & NCLoss \cite{tang2018normalized}      & CVPR'18  & 72.8     & -    \\
    & PSI \cite{xu2021scribble}             & ICCV'21  & 74.9     & -    \\
    & Scribbles + SAM                       &  --      & 75.9     & 76.6 \\
    \midrule[0.8pt]
    \multirow{5}*{Bounding boxes}     
    & WSSL \cite{papandreou2015weakly} & ICCV'15     &  60.6      & 62.2   \\
    & BoxSup \cite{dai2015boxsup}      & ICCV'15     &  62.0      & 64.6   \\
    & SDI \cite{khoreva2017simple}     & CVPR'17     &  69.4      &  -     \\
    & Song \etal \cite{song2019box}    & CVPR'19     &  70.2      &  -     \\
    & BBAM \cite{lee2021bbam}          & CVPR'21     &  73.7      & 73.7   \\
    & Bounding boxes + SAM             &  --         &  76.3      & 75.8   \\ 
    \bottomrule[1pt]
    \end{tabular} }
    \vspace{-5pt}
\end{table}

\section{Experiment Setting}
All the experiments are conducted on PASCAL VOC 2012 dataset, 
which contain 10582/1449/1456 images in the train/val/test set.
We utilize the third mask of SAM's three output masks to 
generate pseudo labels.
Following \cite{jiang2022l2g}, DeepLab-v2 \cite{chen2017deeplab} 
based on ResNet-101 is selected as the default segmentation network, 
whose parameters are initialized using the COCO pre-trained model.
We keep the same training settings with \cite{jiang2022l2g}.

\section{Conclusion}
In this report, we have conducted experiments to explore 
the potential of SAM for generating accurate object masks.
Experiments on PASCAL VOC 2012 dataset demonstrate that 
SAM can serve as a good pseudo-label generator.
In the future, we plan to conduct experiments on 
more complex datasets, such as MS COCO.
Besides, we plan to explore the potential of SAM for 
instance segmentation task.

{
\footnotesize
\bibliographystyle{plain}
\bibliography{lsg}
}

\end{document}